\documentclass[10pt,twocolumn,letterpaper]{article}

\usepackage[pagenumbers]{cvpr} 

\definecolor{cvprblue}{rgb}{0.21,0.49,0.74}
\usepackage[pagebackref,breaklinks,colorlinks,allcolors=cvprblue]{hyperref}
\usepackage{graphicx}
\usepackage{multicol}
\usepackage{multirow}
\usepackage{amsmath}
\usepackage{siunitx}

\usepackage{appendix}
\usepackage{academicons} 
\usepackage{makecell}
\usepackage[table]{xcolor}
\usepackage[linesnumbered,ruled,vlined]{algorithm2e}
\usepackage{pifont}
\usepackage{paralist}
\usepackage{enumitem}
\usepackage{tikz}
\newcommand{\circled}[1]{%
  \tikz[baseline=(char.base)]\node[draw, shape=circle, inner sep=0.2pt, scale=0.9] (char) {#1};%
}

\definecolor{lightgray}{rgb}{0.90, 0.90, 0.90}
\definecolor{lightblue}{rgb}{0.80, 0.90, 0.95}
\definecolor{lighterblue}{rgb}{0.88, 0.94, 0.98}

\def\paperID{3000} 
\def\confName{CVPR}
\def\confYear{2026}

\title{ViSS-R1: Self-Supervised Reinforcement Video Reasoning}

\author{Bo Fang$^{1}$, Yuxin Song$^{2}$, Qiangqiang Wu$^{1}$, Haoyuan Sun$^{2,3}$, Wenhao Wu$^{4}$, Antoni B. Chan$^{1}$\\
$^{1}$ City University of Hong Kong \qquad $^{2}$ Baidu Inc. \\
$^{3}$ Tsinghua University \qquad
$^{4}$ The University of Sydney  \\
}

\begin{document}
\maketitle
\begin{abstract}
Complex video reasoning remains a significant challenge for Multimodal Large Language Models (MLLMs), as current R1-based methodologies often prioritize text-centric reasoning derived from text-based and image-based developments. In video tasks, such  strategies frequently underutilize rich visual information, leading to potential shortcut learning and increased susceptibility to hallucination. 
To foster a more robust, visual-centric video understanding, we start by introducing a 
novel self-supervised reinforcement learning GRPO algorithm (Pretext-GRPO) within the standard R1 pipeline, in which positive rewards are assigned for correctly solving pretext tasks on transformed visual inputs, which makes the model to non-trivially process the visual information.
Building on the effectiveness of Pretext-GRPO, we further propose the ViSS-R1 framework, which streamlines and integrates pretext-task-based self-supervised learning directly into the MLLM's R1 post-training paradigm.
Instead of relying solely on sparse visual cues, our framework compels models to reason about transformed visual input by simultaneously processing both pretext questions (concerning transformations) and true user queries.
This necessitates identifying the applied transformation and reconstructing the original video to formulate accurate final answers.
Comprehensive evaluations on six widely-used video reasoning and understanding benchmarks demonstrate the effectiveness and superiority of our Pretext-GRPO and ViSS-R1 for complex video reasoning.
Our codes and models will be publicly available. 
\end{abstract}
\section{Introduction}
\label{sec:intro}

Beyond fundamental video understanding, complex video reasoning presents new challenges in inferring objects, relationships, events and causality from video content~\cite{zhou2018temporal,yiclevrer,yang2025thinking,zhang2025thinking}.
Following the success of post-training via Reinforcement Learning with Verifiable Reward (RLVR)~\cite{lambert2024tulu3,guo2025deepseekr1} in the large language model domain, recent research has increasingly focused on incentivizing Multimodal Large Language Models (MLLMs) reasoning abilities for complex images~\cite{huang2025visionr1,meng2025mmeureka,zhang2025r1-vl,liu2025visual,tan2025reasonrft,yang2025r1onevision} and videos~\cite{shi2025enhancing,feng2025videor1,wang2025videorft,li2025videochatr1}, by revealing the thinking process in text-based Chain-of-Thought (CoT). 
Despite these early advances, video reasoning with MLLMs remains underexplored.

\begin{figure}
    \centering
    \includegraphics[width=1.0\linewidth]{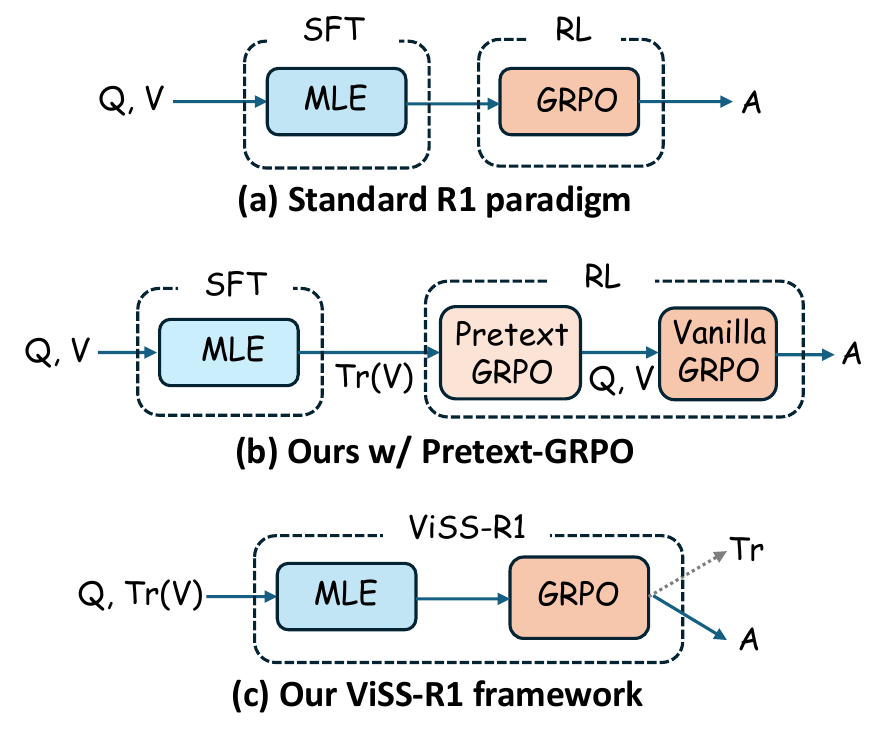}
    \vspace{-6.5mm}
    \caption{\textbf{Overview. }
    \textbf{(a)} Standard R1 paradigm for MLLM video reasoning  consists of a SFT memorization stage followed by RL exploration. 
    \textbf{(b)} We introduce an intermediate Pretext-GRPO stage for visual-centric RL reasoning by learning from self-supervised visual transformations Tr(V).
    \textbf{(c)} Our ViSS-R1 framework fully integrates pretext-task reasoning into the R1-paradigm, where the model takes transformed visual inputs Tr(V) and is designed to simultaneously output the inferred transformation (Tr) and the query's answer (A).
    }
    \label{fig:overview}
    \vspace{-3mm}
\end{figure}

Inspired by image-based MLLM reasoning, recent video reasoning largely employs a similar framework based on the R1 paradigm~\cite{trung2024reft,guo2025deepseekr1}. As illustrated in Fig.~\ref{fig:overview}(a), MLLMs are first subjected to supervised finetuning (SFT) using CoTs, followed by refinement with reinforcement learning (RL) algorithms (e.g., PPO~\cite{schulman2017proximal}, GRPO~\cite{shao2024deepseekmath}).
To enhance temporal reasoning abilities, high-quality video-specific CoT datasets are carefully constructed for SFT training.
While for the RL optimization, methodological advances are  limited to temporal augmentation~\cite{feng2025videor1} or temporal-aligned rewards addition~\cite{li2025videochatr1,wang2025videorft}.

Within these existing video R1 works, dense video information is primarily used as contextual evidence, from which the model extracts sparse cues to support text-based reasoning. 
 (We include an illustrative experiment in appendix.) 
Despite the careful construction of Video CoT datasets, substantial video content remains underutilized, as videos typically contain much more information than can be conveyed by text. 
Consequently, models may exploit shortcuts by focusing on a single frame or snapshot, rather than engaging in deep, comprehensive video reasoning~\cite{yu2025unhackable}. 
Moreover, the predominance of text-centric reasoning in current methodologies, which often overlooks rich visual information, increases the risk of hallucinated generations~\cite{huang2025survey}.
This raises an important question: how can we facilitate MLLM video reasoning from a \textit{visual-centric} perspective?

We propose the \textbf{Vi}deo \textbf{S}elf-\textbf{S}upervised \textbf{R}einforcement framework, termed as \textbf{ViSS-R1}.
To address the challenge of sparse utilization of visual information in videos, we employ pretext-task-based self-supervised learning (SSL)  to enhance visual-centric representation learning for the video. 
SSL has a rich history of leveraging various pretext transformations for representation learning, such as  rotation~\cite{gidaris2018rotnet}, patch shuffling~\cite{noroozi2016jigsaw-puzzle,doersch2015unsupervised}, and inpainting~\cite{pathak2016context} in images, as well as 3D rotation~\cite{jing20183drotnet}, clip shuffling~\cite{xu2019vcop}, and acceleration~\cite{benaim2020speednet,yao2020prp} in videos. The resulting pre-trained models are then transferred to fine-tune specific downstream tasks. 
Within the modern RLVR post-training framework, we utilize pretext SSL as an effective entry point for first extracting and understanding video information, before proceeding to targeted text-based and query-specific reasoning. 
This approach is motivated by the association that visual-centric pretext tasks provides natural reward signals well-suited for RL mechanism, without additional annotations.

In this paper, we propose two methods to introduce SSL into video-based MLLMs: 1) a warm-start pretext-GRPO algorithm based on SSL; 2) an integration of the pretext task into the full training pipeline of the MLLM. 
For the first approach, we introduce a separate Pretext-GRPO stage prior to standard RL training, aiming to provide a \textit{warm start} for the RL policy model, in contrast to the cold start SFT (\cref{fig:overview}(b)). In Pretext-GRPO, visual inputs (including mixed images and videos) are subjected to various transformations, which are then used to train the policy model to identify the specific transformation applied. The optimization algorithm is identical to that of vanilla GRPO, except that the ground-truth labels correspond to specific transformation types.
Subsequently, after applying normal GRPO optimization with real user questions, we demonstrate consistent improvement brought by Pretext-GRPO upon multiple video reasoning and understanding benchmarks.

In the above method,  Pretext-GRPO is decoupled from the standard RL paradigm, which introduces additional training control complexity in practice.
%
Thus, we propose to retain the standard R1 framework, but instead, compel MLLMs to directly reason about transformed videos (\cref{fig:overview}(c)). 
Specifically, the model is simultaneously prompted with both the pretext question and the user's query, which necessitates the model to accurately answer the primary question by correctly identifying the applied transformation first, implicitly recovering the original videos.
This integrated setting is maintained across both the SFT and RL stages: in SFT, the model learns to organize responses for both questions in a predefined format, while in RL, it is jointly optimized with two distinct reward signals.
The complete pipeline constitutes our ViSS-R1.

In summary, our contributions are three-fold:
\begin{compactitem}
    \item We introduce a novel Pretext-GRPO RL algorithm for visual-centric MLLM video reasoning, utilizing self-supervised visual transformations. This annotation-free method can be seamlessly and effectively integrated into existing R1-paradigm frameworks.
    \item We establish the ViSS-R1 framework, which enhances video reasoning by training MLLMs to directly process transformed inputs in both SFT and RL stages.
    \item Comprehensive evaluations on multiple video reasoning and understanding benchmarks demonstrates the effectiveness and superiority of our approaches. 
\end{compactitem}

\section{Related Work}
\label{sec:relate_work}

\begin{figure*}[t]
  \centering
   \includegraphics[width=1.0\linewidth]{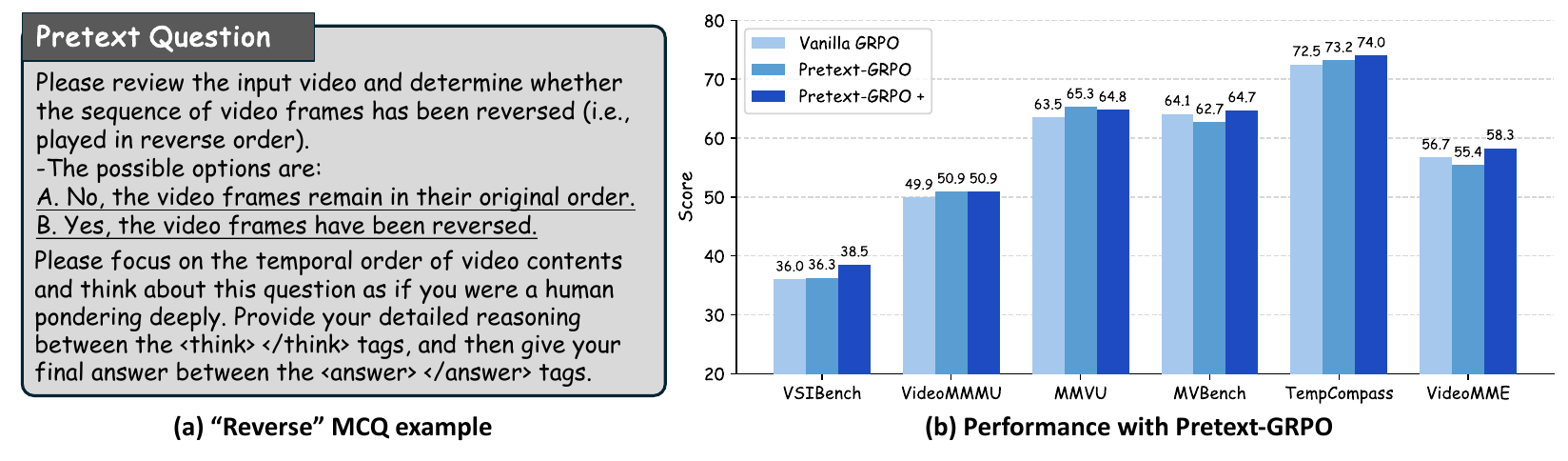}
   \vspace{-8mm}
   \caption{(a) \textbf{Example of a ``Reverse" MCQ pretext question} used in our Pretext-GRPO, where randomly transformed visual inputs are leveraged to construct targeted pretext queries for policy model prompting.
   (b) \textbf{Pretext-GRPO+} denotes a Pretext-GRPO stage followed by vanilla GRPO, which consistently improves performance across multiple video benchmarks. All results are based on 16-frame evaluation.}
   \label{fig:pretext-grpo}
   \vspace{-3mm}
\end{figure*}

\subsection{Multimodal Large Language Model for Video}
\label{subsec:mllm_reason}

\textit{Video understanding} is a fundamental yet challenging task that aims to effectively manage, analyze, and interpret complicated video content~\cite{tang2025videosurvey}. 
Recent advances in large-scale visual instruction tuning led to powerful open-sourced 
MLLMs~\cite{liu2023visual,zhang2024video,li2024llavaonevision,chen2024internvl,wang2024qwen2,bai2025qwen2.5vl}, which have significantly improved video understanding. 
By focusing on video-specific spatio-temporal perception, current video MLLMs~\cite{li2024llamavid,cheng2024videollama2,zhang2024longva,chen2024sharegpt4video,jin2024chatuniv,xu2024pllava,yu2024framevoyager,lin2024vila,chen2024longvila} have achieved notable progress in general tasks such as video question answering and captioning.
Nevertheless, when addressing complex spatio-temporal video reasoning scenarios~\cite{yang2025thinking,hu2025videommmu,zhao2025mmvu}, these models remain inadequate due to  relative lower performance and a lack of thinking abilities.

\textit{Video reasoning} presents new challenges in inferring objects, relationships and causality for visual spatial and temporal intelligence. Inspired by breakthroughs from OpenAI-o1~\cite{jaech2024openai-o1} and DeepSeek-R1~\cite{guo2025deepseekr1} in lifting the reasoning abilities of LLMs through RL, 
numerous studies~\cite{huang2025visionr1,yang2025r1onevision,zhang2025r1-vl,meng2025mmeureka,deng2025openvlthinker,sun2025reinforcement} have adopted the R1 paradigm to incentivize the visual reasoning abilities of MLLMs.
Specifically for videos, Video-R1~\cite{feng2025videor1} adapts the R1 pipeline to the video domain and introduces shuffling augmentation to improve temporal reasoning.
VideoRFT~\cite{wang2025videorft} further incorporates a semantic-consistency reward to better align textual reasoning with visual information.
Given the temporal structure and rich content of videos, researchers have focused on designing and integrating multiple types of temporal-related rewards~\cite{li2025temporalrlt,li2025videochatr1} into the original system.
Despite these advances, current video reasoning approaches leverage only sparse visual cues for text-centric thinking, leaving substantial dense video information underutilized (\S\ref{sec:intro}).
%
In contrast to previous works, we propose novel self-supervised RL algorithms that enable explicit visual processing, thereby enhancing subsequent deep video reasoning by utilizing more comprehensive video information.

\subsection{Self-Supervised Representation Learning}
\label{subsec:ssl}
%

Image-oriented self-supervised learning (SSL) has primarily employed transformation-based pretext tasks for upstream pretraining, such as solving jigsaws~\cite{noroozi2016jigsaw-puzzle,chen2023jigsawvit}, predicting rotation angles~\cite{gidaris2018rotnet}, localizing patches~\cite{doersch2015unsupervised}, etc., which have subsequently inspired analogous video SSL tasks like 3D rotation~\cite{jing20183drotnet}, space-time cubic puzzles~\cite{kim2019spacetimepuzzle}. 
Furthermore, various temporal-specific tasks, including sorting frames~\cite{lee2017unsupervised,misra2016shuffle} or clips~\cite{xu2019vcop}, perceiving the arrow of time~\cite{wei2018arrow}, discriminating among multiple temporal samplings~\cite{luo2020VCP,jenni2020video}, have been proposed for vision-only video models on extracting spatiotemporal representations for classification/recognition tasks.
These effective designs motivate us to integrate such pretext transformations into the current RLVR framework for intermediate, visual-centric reasoning steps.
Although contrastive learning algorithms~\cite{he2020moco,chen2020SimCLR,grill2020BYOL} and reconstruction-based methods~\cite{he2022mae,tong2022videomae,wang2023videomaev2} have advanced SSL for large-scale vision foundation models, their complex training frameworks hinder direct adoption in standard RL settings.

Two concurrent works, Jigsaw-R1~\cite{wang2025jigsawr1} and VisualJigsaw~\cite{wu2025visualjigsaw}, pursue similar self-supervised RL approaches by post-training MLLMs on the image jigsaw task. 
In contrast, we focus more on video reasoning and incorporate a broader range of pretext transformations beyond single jigsaw objective. Furthermore, we systematically investigate a streamlined and effective integration of visual-centric SSL within the current R1 framework, moving beyond simple combinations of pretext RL tasks.

\section{Method}
\label{sec:method}


\begin{figure*}[t]
  \centering
   \includegraphics[width=1.0\linewidth]{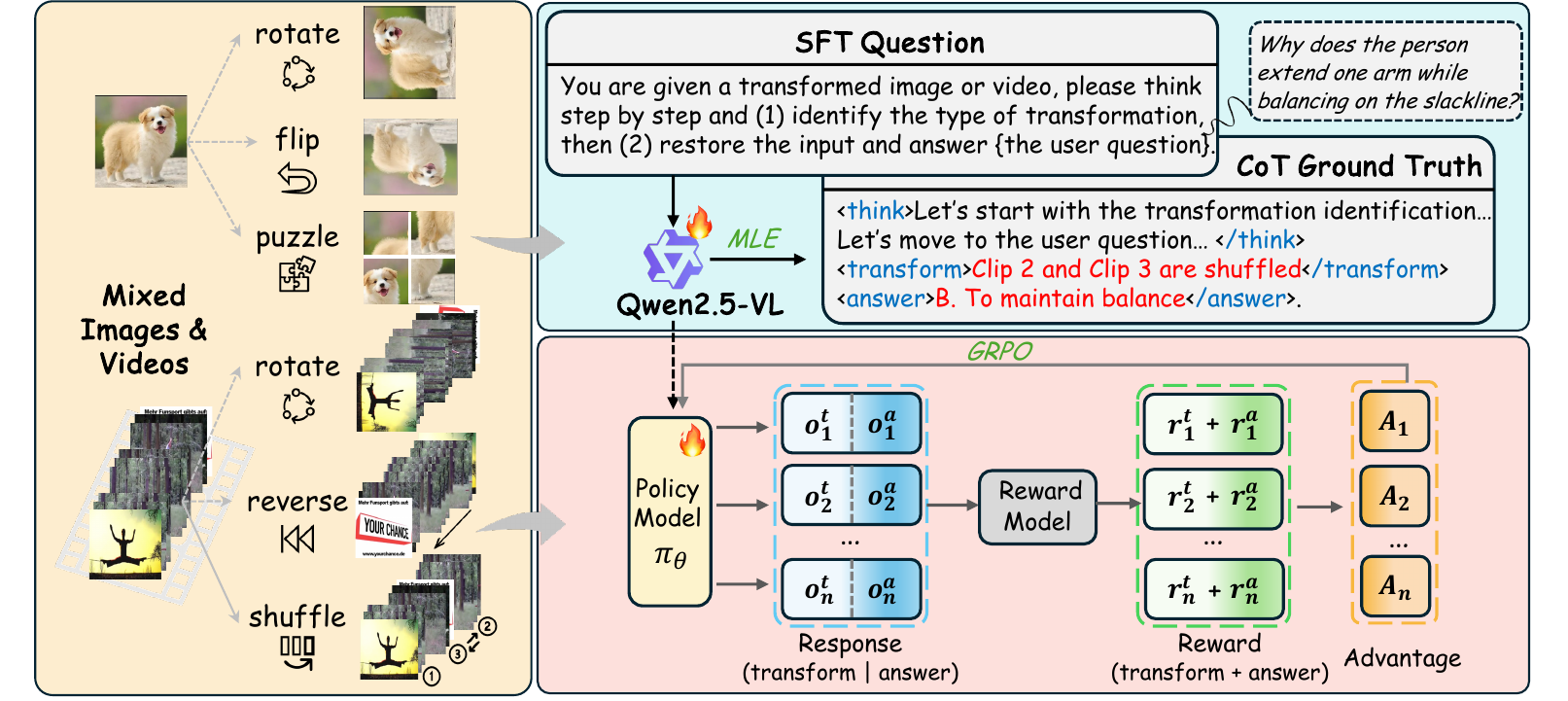}
   \vspace{-6.5mm}
   \caption{\textbf{ViSS-R1 framework.}
   Mixed images and videos are randomly augmented with SSL transformations for both SFT and RL reasoning. The models are required to simultaneously address pretext questions regarding the applied transformations and answer real user queries.
    ViSS-R1 additionally learns a \texttt{<transform>} tag in SFT to encapsulate pretext identification results, which provides structural organization for answers and facilitates reward manipulation during RL exploration.
   }
   \label{fig:method}
   \vspace{-3mm}
\end{figure*}

\subsection{Vision-Oriented Pretext-GRPO}
\label{subsec:method-sub-1}

\noindent
\textbf{Preliminaries.} 
Recent studies aiming to incentivize the reasoning capabilities of MLLMs typically adopt a paradigm consisting of SFT cold start followed by RL refinement.
With carefully curated long CoT, the SFT stage enables the model to learn the semantics of specialized tag tokens and to autoregressively generate structured outputs through thinking before producing final answers, e.g., \texttt{<think>$\cdots$</think><answer>$\cdots$</answer>}.
During the RL stage, a commonly used algorithm is Group Relative Policy Optimization (GRPO)~\cite{shao2024deepseekmath}, which is a variant of Proximal Policy Optimization (PPO)~\cite{schulman2017proximal}, that eliminates the value model and significantly lowers the cost of training resources. 
Given an SFT initialized policy model $\pi_{\theta}$ and a reference model $\pi_{\mathrm{ref}}$, GRPO first generates a set of responses $\{o_1, o_2, \cdots, o_G\}$ through policy sampling. A group-based reward function then computes the corresponding rewards $\{r_1, r_2, \cdots, r_G\}$ against the ground truth, which are subsequently used to estimate the advantage $A_i$ for each response relative to the group:
\begin{align}
    A_i = \tfrac{r_i-\mathrm{mean}(\{r_1, r_2, \cdots, r_G\})}{\mathrm{std}(\{r_1, r_2, \cdots, r_G\})}.
\end{align}
Moreover, GRPO employs a clipped objective with a KL penalty term to stabilize the training process:
\begin{equation}
\begin{aligned}
    \mathcal{J}_{\mathrm{GRPO}} =& \mathbb{E}_{q,\{o_i\}}
    \Big[
        \tfrac{1}{G}\sum\nolimits_{i=1}^{G}
        \big( \min\big(\tfrac{\pi_{\theta}(o_i|q)}{\pi_{\theta_{\mathrm{old}}}(o_i|q)}A_i, \\
         & \mathrm{clip}(\tfrac{\pi_{\theta}(o_i|q)}{\pi_{\theta_{\mathrm{old}}}(o_i|q)}, 1-\epsilon, 1+\epsilon)A_i \big)- \\
         & \beta \mathbb{D}_{\mathrm{KL}}(\pi_{\theta} \parallel \pi_{\mathrm{ref}}) \big) 
    \Big].    
\end{aligned}
\label{eq:grpo}
\end{equation}

\noindent
\textbf{Pretext-GRPO.}
Beyond trivial memorization in SFT~\cite{chusft}, GRPO encourages models to reason creatively along diverse trajectories by rewarding those that lead to correct final outcomes. 
However, despite its advantages, vanilla GRPO in MLLMs often overlooks rich visual information when answering localized questions, particularly in videos~\cite{feng2025videor1,wang2025videorft} (see \S\ref{sec:intro}).
To address this limitation, we propose Pretext-GRPO that conducts RL on self-supervised transformations, enabling the policy model to effectively leverage and explore rich, vision-oriented information.

Given a training image or video $V$, we do not use its real user question $Q$ for Pretext-GRPO. 
Instead, a handcrafted pretext question $Q_p$, together with the corresponding transformed visual input $\mathrm{Tr}(V)$, are inputs to the policy model for RL.
To facilitate straightforward reward assignment, all $Q_p$ are formulated as multiple-choice questions (MCQs).
Specifically for images, we consider three types of self-supervised transformations (SSTs), enabling the model to learn spatial semantics by solving spatial-related tasks:
\setdefaultleftmargin{1em}{2.2em}{1.87em}{1.7em}{1em}{1em}
\begin{compactitem}
\item \textit{Rotate}~\cite{gidaris2018rotnet}: The image is randomly rotated by 0$^{\circ}$, 90$^{\circ}$, 180$^{\circ}$, or 270$^{\circ}$, resulting in a 4-option MCQ.
\item \textit{Flip}: The image is randomly flipped vertically or horizontally, including no flip, forming a 3-option MCQ.

\item \textit{Puzzle}~\cite{noroozi2016jigsaw-puzzle,doersch2015unsupervised}: The image is evenly divided into four patches by splitting it along both the horizontal and vertical axes; two patches are then randomly selected and their positions swapped, resulting in a 6-option MCQ.
\end{compactitem}
\noindent
Similarly, three SSTs are randomly applied to videos to capture spatio-temporal representations:
\begin{compactitem}
\item \textit{3D Rotate}~\cite{jing20183drotnet}: All frames in a video are randomly rotated by the same degree, resulting in a 4-option MCQ.
\item \textit{Reverse}~\cite{wei2018arrow}: The video is either presented in its original direction or reversed, forming a binary MCQ.
\item \textit{Shuffle}~\cite{xu2019vcop}: The video is divided into 4 consecutive clips, of which 2 are shuffled, 
yielding a 6-option MCQ.
\end{compactitem}
\noindent
We present an example of a ``Reverse" pretext question in \cref{fig:pretext-grpo}(a), with full prompts of these questions in the appendix.

The above SSTs with MCQ formulations provide ``free" supervision that can be interpreted as rewards during RL. 
As in vanilla GRPO, our Pretext-GRPO assigns a reward of 1 if the pretext question is correctly answered, and 0 otherwise. 
The policy model $\pi_{\theta}$ is then updated using the GRPO objective in (\ref{eq:grpo}), based on the pretext question $Q_p$ and transformed input $\mathrm{Tr}(V)$, i.e., $\frac{\pi_{\theta}(o_i|\mathrm{Tr}(V), Q_p)}{\pi_{\theta_{\mathrm{old}}}(o_i|\mathrm{Tr}(V), Q_p)}$.
Unlike cold-start SFT, which employs token-level regression to memorize and imitate the reasoning path~\cite{chusft}, Pretext-GRPO provides a \textit{warm start} for the policy model by enabling it to self-examine internal visual content and semantics prior to standard RL. 
The effectiveness of Pretext-GRPO is demonstrated in \cref{fig:pretext-grpo}(b) across multiple video reasoning and understanding benchmarks (more in \S\ref{subsec:ex_setup}). 
When combining vanilla GRPO on true user questions and raw visual input after Pretext-GRPO (denoted as \textbf{Pretext-GRPO+}), we observe consistent improvements across all benchmarks.

%

\subsection{Vision-Integrated ViSS-R1}
\label{subsec:method-sub-2}
Initiating RL with a separate Pretext-GRPO stage helps stabilize standard RL training by warm-starting the policy model to recognize spatiotemporal visual transformations.
However, splitting RL into two distinct stages introduces additional training control complexity because the two stages optimize different objectives. 
This motivates us to integrate SSL into a single-stage RL procedure that directly answers user questions based on transformed image/video inputs.
We therefore introduce the ViSS-R1 framework (see \cref{fig:method}).

\noindent 
\textbf{SFT with transform-tagged CoT.}
Base MLLMs are generally unable to answer two questions concurrently in a structured manner, even with explicit instructions.
To enable the categorization of two answers from single-round outputs, 
we teach models to learn generating responses in a predefined structure using SFT with transform-tagged CoTs. 
Building on previous CoT construction, we introduce an additional \texttt{<transform></transform>} tag to encapsulate the results of the final predicted transformation.
Following Video-R1~\cite{feng2025videor1}, our ground-truth CoT rationales are reconstructed by distilling answers from the advanced Qwen2.5-VL-72B model~\cite{bai2025qwen2.5vl} on randomly transformed images and videos (\cref{subsec:method-sub-1}).
%
During SFT, we apply the \textit{identical} transformation to each input as used for prompting the 72B model and train the base model using the MLE loss, as illustrated in Fig.~\ref{fig:method} (top right).
This process enables the base model to learn to reason and respond to both questions within distinct tags, which correspond to the two rewards in the subsequent RL stage.

\noindent
\textbf{RL reasoning on transformed inputs.}
Given transformed visual inputs $\mathrm{Tr}(V)$, a pretext question $Q_p$ and a real user query $Q$, our transformation-acquainted SFT model is required to solve the self-supervised task (implicitly restoring the inputs) and answering real questions via GRPO algorithm.
Compared to vanilla GRPO, the key distinction of ViSS-R1 is that two types of responses ($o_i^t$ for transformation identification result and $o_i^q$ for user question answers) are sampled in a single generation round.
The importance sampling ratio between the new policy $\pi_{\theta}$ and the old policy $\pi_{\theta_{\mathrm{old}}}$ in (\ref{eq:grpo}) can be rewritten as:
\begin{align}
    \tfrac{\pi_{\theta}\big(o_i^t, o_i^a|\mathrm{Tr}(V), Q_p, Q\big)}
    {\pi_{\theta_{\mathrm{old}}}\big(o_i^t, o_i^a|\mathrm{Tr}(V), Q_p, Q\big)}.
\end{align}


\noindent
\textbf{Reward manipulation.} 
ViSS-R1 incorporates three types of rewards for RL: 
$\circled{1}$ \textbf{Transformation reward $R_t$}: The initial task is to identify the transformation applied to images or videos,
and a reward of 0.5 is assigned if the pretext question $Q_p$ is correctly answered, and 0 otherwise.
$\circled{2}$ \textbf{Accuracy reward $R_a$}: Following previous works~\cite{feng2025videor1,wang2025videorft}, we adopt task-specific accuracy metrics, including Exact Match for multiple-choice and numerical questions, ROUGE for open-ended generation, Word Error Rate for OCR tasks, and a scaled relative accuracy for regression problems.
$\circled{3}$ \textbf{Format reward $R_f$}: The model’s output is required to follow a predefined structure: the thinking process must be enclosed in \texttt{<think>$\cdots$</think>} tags, transformation prediction in \texttt{<transform>$\cdots$</transform>}, and answers to user questions in \texttt{<answer>$\cdots$</answer>}. 
The overall reward for a transformed sample is 
$R=R_t+R_a+R_f$.

\section{Experiment}
\label{sec:experiment}

\subsection{Experimental Setup}
\label{subsec:ex_setup}

\textbf{Benchmarks and metric.} Following recent work~\cite{feng2025videor1,wang2025videorft,zhang2025tinyllavavideor1}, we evaluate our model on three general video reasoning benchmarks including VSI-Bench~\cite{yang2025thinking}, VideoMMMU~\cite{hu2025videommmu}, MMVU~\cite{zhao2025mmvu}, and three general video understanding benchmarks: MVBench~\cite{li2024mvbench}, TempCompass~\cite{liu2024tempcompass} and VideoMME~\cite{fu2025videomme} (w/o subtitles).
For MMVU, only the multiple-choice subset are used for evaluation. 
We report average accuracy (Acc) for all above tasks.

\noindent
\textbf{Model training.}
We use Qwen2.5-VL-7B~\cite{bai2025qwen2.5vl} as the base model, and our training dataset is derived from Video-R1~\cite{feng2025videor1} (i.e., Video-R1-CoT-165k and Video-R1-260k).
For efficiency, we sample 32 frames from each video and limit the maximum resolution of each frame to 128$\times$28$\times$28 during training.
Pretext-GRPO is trained for 500 steps, followed by vanilla GRPO for 1K steps, denoted as Pretext-GRPO+.
In integrated ViSS-R1, we reprompt Qwen2.5-VL-72B for transformation-acquainted CoT construction and SFT training. 
All models are trained on 8 NVIDIA A800 (80G) GPUs. Our codebase is built on Open-R1~\cite{faceopen}.

\noindent
\textbf{Inference.}
For inference, we sample 32 frames and increase the resolution to 256$\times$28$\times$28 to enhance performance, following works~\cite{feng2025videor1,wang2025videorft}. The decoding configuration follows the Qwen2.5-VL demo, with top\_p=0.001 and temperature=0.01.
During ViSS-R1 inference, pretext questions are removed from the testing prompt, and models are required to reason on untransformed, raw videos.

\begin{table*}
    \caption{\textbf{Performance comparisons.} VSI refers to VSI-Bench, and TempC refers to the TempCompass benchmark. $^*$ indicates results reported with higher input resolutions and more frames (i.e., 768$\times$28$\times$28 and 768 sampled frames), while our results use 256$\times$28$\times$28 and 32 frames.
    \textbf{Boldface} and \underline{underline} indicate the best and second-best results, respectively.}
    \vspace{-2mm}
    \label{tab:sota}
    \small
    \centering
    \begin{tabular}{@{}lr|ccc|ccc@{}}
    \toprule
    \multirow{2}{*}{\textbf{Model}} & \multirow{2}{*}{\textbf{Venue}} & \multicolumn{3}{c}{\textbf{Video Reasoning}}  \vline & \multicolumn{3}{c}{\textbf{Video Understanding}} \\
    & & VSI. & VideoMMMU & MMVU &  MVBench & TempC. & VideoMME  \\
    \hline
    \rowcolor{lightgray}
    \multicolumn{8}{c}{\textit{Proprietary Models}}\\
    GPT-4o~\cite{hurst2024gpt4o} &  & 34.0 & 61.2 & 75.4 & - & - & 71.9  \\
    \hline
    \rowcolor{lightgray}
    \multicolumn{8}{c}{\textit{Open-Source Models}}\\
    LLaMA-VID~\cite{li2024llamavid} & ECCV 24 & - & - & - & 41.9 & 45.6 & - \\
    ShareGPT4Video~\cite{chen2024sharegpt4video} & NeurIPS 24 & - & - & - & 51.2 & - & 39.9 \\
    VideoLLaMA2~\cite{cheng2024videollama2} & arXiv 24.06 & - & - & 44.8 & 54.6 & - & 47.9 \\
    LongVA-7B~\cite{zhang2024longva} & TMLR 24 & 29.2 & 23.9 & - & - & 56.9 & 52.6 \\
    VILA-1.5-8B~\cite{lin2024vila} & CVPR 24 & 28.9 & 20.8 & - & - & 58.8 & - \\
    Video-UTR-7B~\cite{yu2025unhackable} & ICLR 25 & - & - & - & 58.8 & 59.7 & 52.6 \\
    mPLUG-Owl3-8B~\cite{ye2024mplugowl3} & ICLR 25 & - & - & - & 54.5 & - & 53.5 \\
    LLaVA-OneVision-7B~\cite{li2024llavaonevision} & TMLR 24 & 32.4 & 33.8 & 49.2 & 56.7 & - & 58.2 \\
    Qwen2.5-VL-7B~\cite{bai2025qwen2.5vl} & arXiv 25.02 & 30.1 & 48.1 & 60.0 & 59.0 & 72.6 & 56.6 \\
    \hline
    \rowcolor{lightgray}
    \multicolumn{8}{c}{\textit{R1-based Models}}\\
    Video-R1~\cite{feng2025videor1} & NeurIPS 25 & 35.8 & \underline{52.3} & 63.8 & 63.9 & 73.2 & 59.3 \\
    TinyLLaVA-Video-R1~\cite{zhang2025tinyllavavideor1} & arXiv 25.04 & - & - & 46.9 & - & 49.5 & 46.6 \\
    VideoChat-R1~\cite{li2025videochatr1} & arXiv 25.04 & - & - & - & {67.9}$^{*}$ & - & - \\
    Temporal-RLT~\cite{li2025temporalrlt} & arXiv 25.06 & - & - & 65.0 & \textbf{68.1} & 73.3 & 57.6 \\
    VideoRFT~\cite{wang2025videorft} & NeurIPS 25 & 36.8 & 51.1 & \textbf{68.5} & 62.1 & 73.7 & 59.8 \\
    \textbf{Pretext-GRPO+ (Ours)} & & \textbf{39.2} & \textbf{53.9} & 65.3 & \underline{66.0} & \underline{73.9} & \underline{60.1} \\
    \textbf{ViSS-R1 (Ours)} & & \underline{37.3} & 51.7 & \underline{66.1} & 65.6 & \textbf{75.3} & \textbf{60.5} \\
    \bottomrule
    \end{tabular}
    \vspace{-3mm}
\end{table*}

\subsection{Comparisons with Previous Methods}
\label{subsec:sota}
We present comprehensive comparisons with previous methods in \cref{tab:sota}, covering proprietary model (i.e., GPT-4o~\cite{hurst2024gpt4o}), open-source MLLMs (e.g., LLaMA-VID~\cite{li2024llamavid}, VILA-1.5~\cite{lin2024vila}, LLaVA-OneVision~\cite{li2024llavaonevision}, etc.), and recent R1-based models (e.g., Video-R1~\cite{feng2025videor1}, VideoRFT~\cite{wang2025videorft}, Temporal-RLT~\cite{li2025temporalrlt}).
Notably, our proposed methods, \textbf{Pretext-GRPO+} and \textbf{ViSS-R1}, achieve state-of-the-art performance on 4 out of 6 video reasoning and understanding benchmarks (\textbf{39.2}\% on VSI-Bench, \textbf{53.9}\% on VideoMMMU, \textbf{75.3}\% on TempCompass, and \textbf{60.5}\% on VideoMME). These improvements demonstrate the effectiveness of our self-supervised RL strategy in incentivizing the video reasoning capabilities of MLLMs.

Additionally, compared to the initial baseline Qwen2.5-VL, ViSS-R1 achieves consistent and significant improvements of \textbf{+7.2\%} on VSI-Bench, \textbf{+4.3\%} on VideoMMMU, and \textbf{+6.1\%} on MMVU on reasoning-specific benchmarks, highlighting the value of post-training techniques for unlocking video reasoning abilities. 
Relative to another baseline, Video-R1 (since we use identical image and video data), Pretext-GRPO+ yields \textbf{+3.4\%} on VSI-Bench and \textbf{+1.6\%} on VideoMMMU, while ViSS-R1 achieves \textbf{+2.1\%} on TempCompass and \textbf{+1.2\%} on VideoMME. 
For small-scale MMVU ($\sim$600 videos), performance can be sensitive to a few additional correct or incorrect answers.
On MVBench, the leading results from VideoChat-R1 and Temporal-RLT are primarily due to their temporal grounding alignment.
Nevertheless, our models deliver competitive results on these benchmarks.
Overall, the superior performance demonstrates the strong potential of integrating self-supervised RL into existing R1-based methods for visual-centric complement.

\subsection{Ablation Study}
\label{subsec:ablation}

\begin{table*}
    \caption{\textbf{Ablation studies of ViSS-R1.}
    Our full model is jointly trained on mixed images and videos, employing a sequential SFT and RL training pipeline. 
    RL supervision is provided via three types of rewards (transformation, accuracy, and format rewards), combined as $R=R_f+R_t+R_a$.
    VSI refers to VSI-Bench, and TempC refers to the TempCompass benchmark. 
    \textbf{Boldface} indicate the best results.}
     \vspace{-2mm}
     \small
    \label{tab:main_ablation}
    \centering
    \begin{tabular}{@{}ll|ccc|ccc@{}}
    \toprule
    \multirow{2}{*}{\textbf{Config}} & \multirow{2}{*}{\textbf{Model}} & \multicolumn{3}{c}{\textbf{Video Reasoning}} \vline &  \multicolumn{3}{c}{\textbf{Video Understanding}} \\
     & & VSI. & VideoMMMU & MMVU & MVBench & TempC. & VideoMME  \\
     \hline
    \rowcolor{lightgray}
    \multicolumn{8}{c}{\textit{Training Data}} \\
    A.1 & Image only & 34.8 & 49.1 & 65.6 & 63.9 & 75.0 & 57.8 \\
    A.2 & Video only & 37.0 & 51.2 & 65.3 & 64.7 & 75.2 & 59.1 \\
    \hline
    \rowcolor{lightgray}
    \multicolumn{8}{c}{\textit{Training Paradigm}} \\
    B.1 & SFT only & 31.7 & 46.4 & 63.0 & 60.9 & 70.6 & 54.7 \\
    B.2 & RL only & 24.0 & 51.4 & \textbf{66.6} & \textbf{65.6} & 73.5 & 59.3 \\
    \hline
    \rowcolor{lightgray}
    \multicolumn{8}{c}{\textit{Reward Modeling}} \\
    C.1 & $R = R_f+R_t$ & 32.0 & 48.7 & 64.5 & 62.3 & 74.3 & 56.0 \\
    C.2 & $R = R_f+R_a$ & 35.8 & 51.6 & 63.2 & 63.6 & 73.9 & 58.0 \\
    \hline
    \rowcolor{lightgray}
    \multicolumn{8}{c}{\textit{Full Model}} \\
    D & \textbf{ViSS-R1 } & \textbf{37.3} & \textbf{51.7} & 66.1 & \textbf{65.6} & \textbf{75.3} & \textbf{60.5} \\
    \bottomrule
    \end{tabular}
\end{table*}

\noindent
\textbf{Training data.}
Our integrated ViSS-R1 framework is trained on mixed images and videos from Video-R1, utilizing 6 types of self-supervised transformations.
We do not ablate each transformation individually, as all pretext tasks have been verified effective in previous literature (\S\ref{subsec:ssl}), and several have already been successfully incorporated into current R1 paradigms~\cite{wu2025visualjigsaw,wang2025jigsawr1}.
In \cref{tab:main_ablation}, we divide the training sources into \textit{Image only} (A.1) and \textit{Video only} (A.2) to evaluate the effectiveness of image-based (rotate, flip, and puzzle) and video-based (rotate, reverse, and shuffle) transformations, separately.
Results indicate that image data, combined with 2D pretext tasks, achieves reasonable performance on most video benchmarks, providing sufficient spatial knowledge for basic video understanding.
Meanwhile, video-specific SSR1 (A.2) consistently outperforms its image-based counterpart, underscoring the critical role of spatiotemporal representation learning for video domain reasoning and understanding.

\noindent
\textbf{Training paradigm.}
We first evaluate the effectiveness of Pretext-GRPO as an intermediate and independent RL stage in \cref{fig:pretext-grpo}(b), where all results are reported using 16 frames for simplicity.
Although not using real questions, Pretext-GRPO alone achieves notable results on reasoning benchmarks such as VideoMMMU (50.9\%) and MMVU (65.3\%), supporting our primary motivation of acquiring vision-centric knowledge prior to localized answering. 
Pretext-GRPO+, which integrates vanilla GRPO with user questions, further enhances and stabilizes performance on more general video understanding tasks.

Moreover, we analyze the sequential SFT and RL training paradigm in ViSS-R1 as shown in \cref{tab:main_ablation}. 
\textit{SFT only} (B.1) refers to training exclusively with supervised fine-tuning on our transformation-reprompt CoT annotations. In this stage, the model learns to follow specific instructions and produce structured responses; however, it lacks evident reasoning abilities, and may suffer from the overfitting problem.
In contrast, \textit{RL only} (B.2), despite omitting SFT initialization, demonstrates strong reasoning capabilities on most benchmarks. 
Notably, the RL-only model achieves significantly lower performance (24.0\%) on VSI-Bench, which contains many challenging regression-type tasks. We observe that the RL-only model struggles to generate well-formatted predictions for such tasks, underscoring the necessity of an SFT cold start.
By adopting standard R1 paradigm (SFT followed by RL), our ViSS-R1 achieves robust and generalizable improvements.

\noindent
\textbf{Reward modeling.}
In \cref{tab:main_ablation} (C.1, C.2), we ablate the impact of reward design, with format reward $R_f$ fixed as an anchor following previous literature~\cite{feng2025videor1,wang2025videorft}.
As demonstrated in Pretext-GRPO, reinforcement learning with only self-supervised transformation identification can yield solid improvements over its SFT model (C.1 \textit{vs.} B.1), particularly on MMVU (64.5\%) and TempCompass (74.3\%).
Furthermore, when jointly trained with accuracy reward $R_a$ on real questions (D), we achieve a comprehensively enhanced and superior model, ViSS-R1.

\begin{figure}[]
  \centering
  \vspace{-3mm}
   \includegraphics[width=1.0\linewidth]{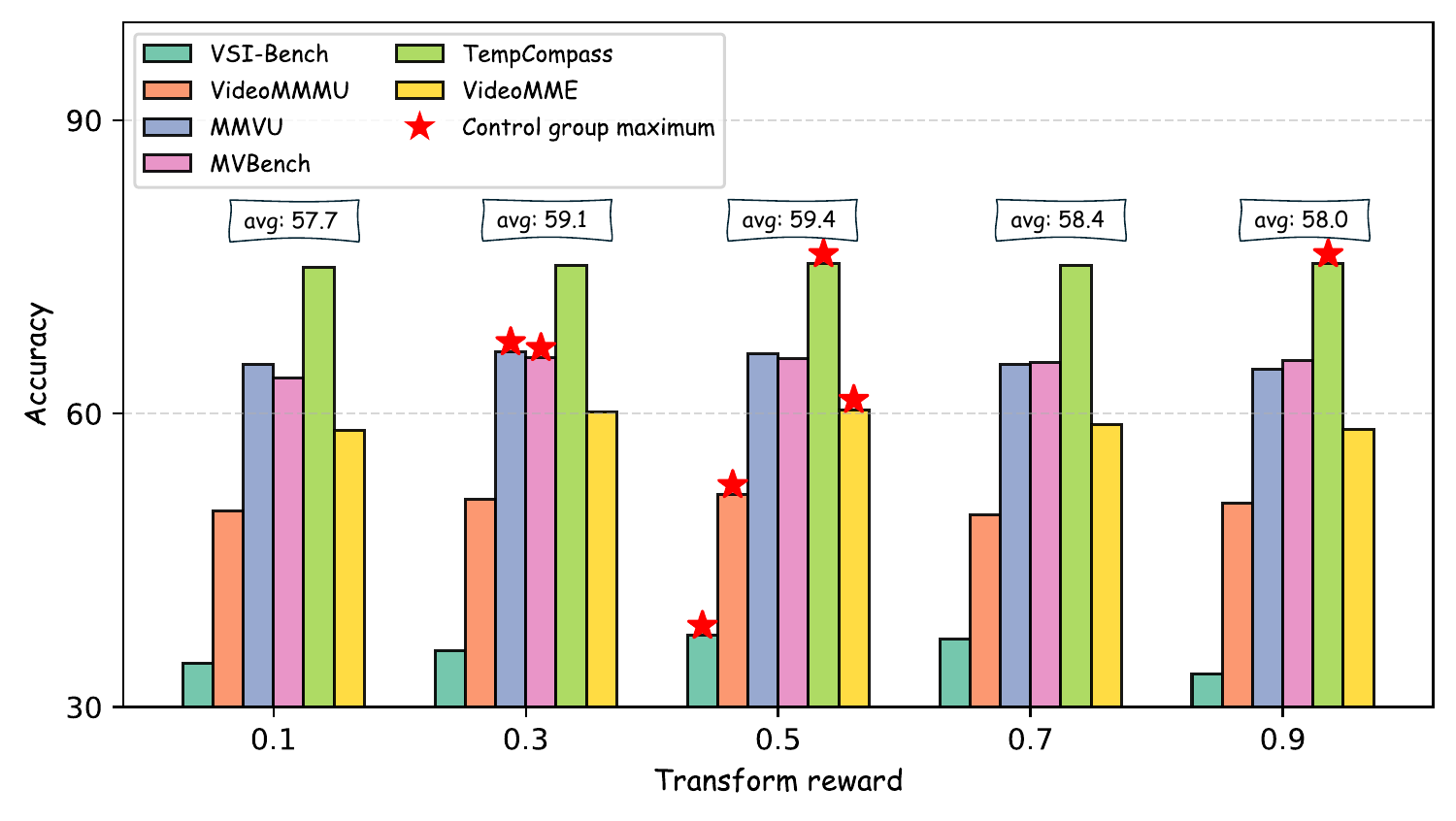}
   \vspace{-8mm}
   \caption{\textbf{Impact of training reward $R_t$.} }
   \vspace{-4mm}
   \label{fig:trans_reward}
\end{figure}

\begin{table}
    \vspace{-3mm}
    \caption{\textbf{Comparisons on transformed videos.} 
    $V$ denotes raw videos, while Tr($V$) represents transformed videos with rotation, reversal, or shuffling. Performance difference $\Delta$ is also reported.}
     \vspace{-2.5mm}
    \label{tab:aug_ablate}
    \centering
    \resizebox{0.47\textwidth}{!}{
    \begin{tabular}{@{}l|ccc|ccc@{}}
    \toprule
    \multirow{2}{*}{\textbf{Model}} & \multicolumn{3}{c}{\textbf{VSI-Bench}} \vline & \multicolumn{3}{c}{\textbf{VideoMME}}\\
    & $V$ & Tr($V$) & $\Delta$  &  $V$ & Tr($V$) & $\Delta$ \\
    \midrule
    Video-R1~\cite{feng2025videor1} & 35.8 & 32.6 & -3.2 & 59.3 & 55.9 & -3.4 \\
    VideoRFT~\cite{wang2025videorft} & 36.8 & 30.8 & \textcolor{red}{-6.0} & 59.8 & 54.6 & \textcolor{red}{-5.2} \\
    \textbf{Pretext-GRPO+ (Ours)} & \textbf{39.2} & \textbf{38.1} & \textcolor{green}{-1.1} & 60.1 & 58.3 & -1.8 \\
    \textbf{ViSS-R1 (Ours)} & 37.3 & 34.8 & -2.5 & \textbf{60.5} & \textbf{59.3} & \textcolor{green}{-1.2} \\
    \bottomrule
    \end{tabular}
    }
    \vspace{-4mm}
\end{table}

\begin{figure*}[t]
  \centering
   \includegraphics[width=0.98\linewidth]{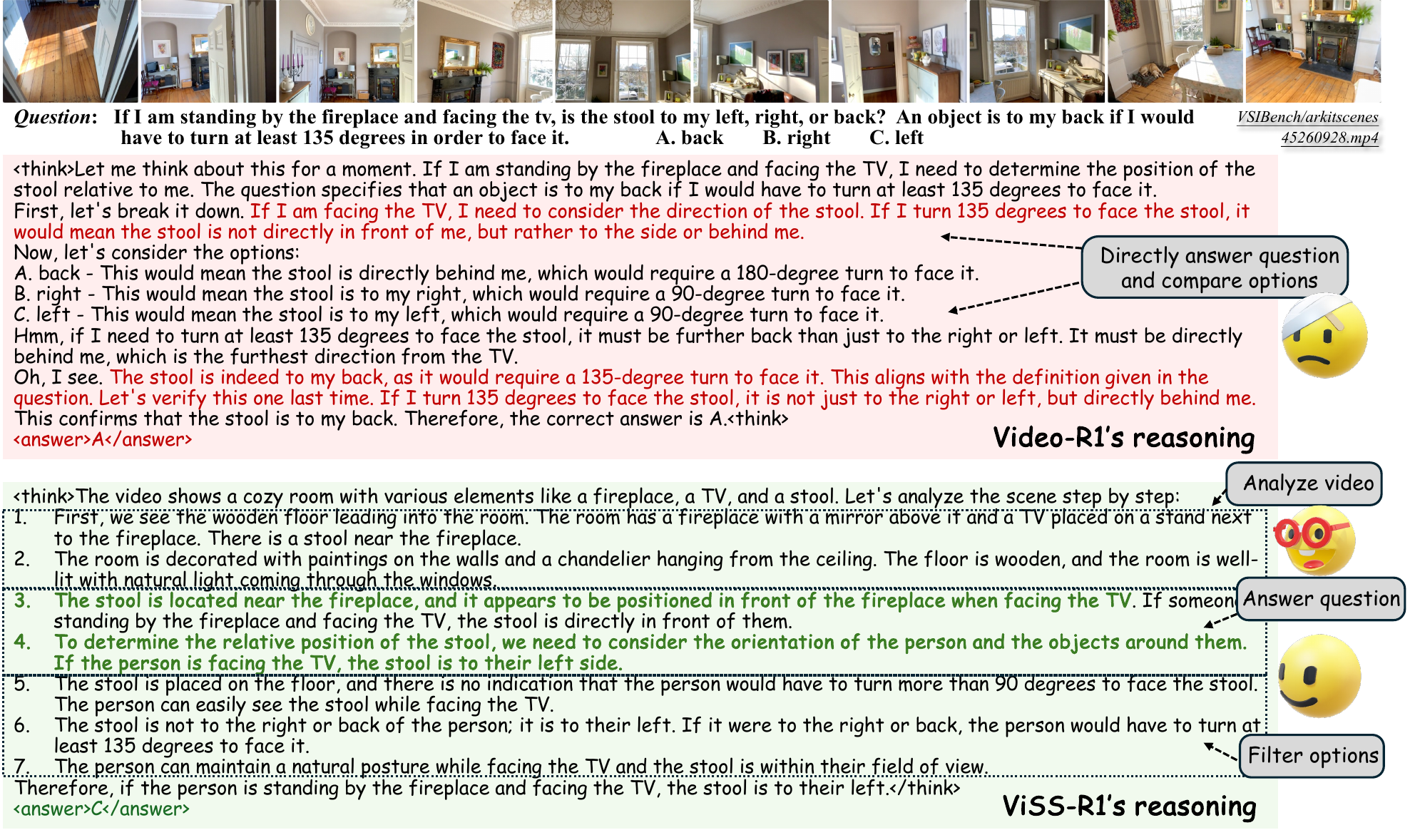}
   \vspace{-4mm}
   \caption{\textbf{Qualitative results.} 
   For ViSS-R1, we remove the pretext question in prompts and use \textbf{un}transformed videos for inference.
   }
   \label{fig:vis}
   \vspace{-3mm}
\end{figure*}

\begin{figure}[]
  \centering
  \vspace{-5mm}
   \includegraphics[width=1.0\linewidth]{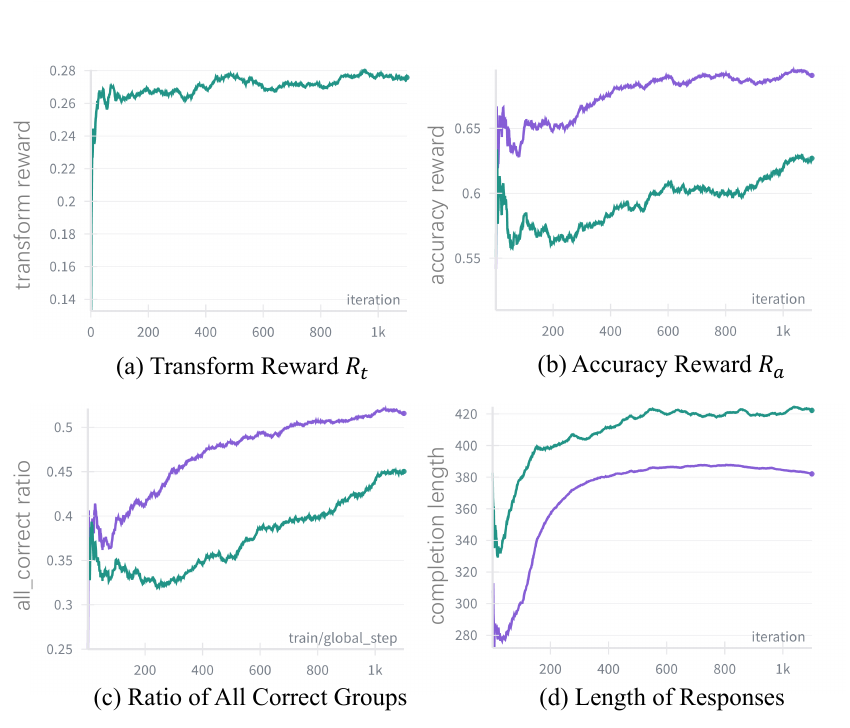}
   \vspace{-8mm}
   \caption{\textbf{Training curves.} 
   \textcolor{violet}{Purple} lines represent results from vanilla GRPO, while \textcolor{green}{green} lines correspond to our ViSS-R1. 
   }
   \vspace{-6mm}
   \label{fig:train_curve}
\end{figure}

\noindent
\textbf{Scaling transformation reward $R_t$.}
Intuitively, pretext tasks should be assigned less importance than real questions when jointly optimized in ViSS-R1.
Here, we analyze the impact of transformation reward $R_t$ while keeping the correctly answered accuracy reward $R_a$ as 1.0 (\cref{fig:trans_reward}).
Across all control group experiments, setting $R_t=0.5$ yields the highest average score of 59.4\% and achieves the best results in 4 out of 6 video benchmarks.
%
Increasing $R_t$ (e.g., to 0.9) leads to a clearly drop in performance, as the model receives a similar reward for answering pretext and real questions, which diminishes its ability to distinguish between them and ultimately impairs its capability.

\noindent
\textbf{Comparisons on transformed videos.}
We evaluate various models on \textit{transformed} videos using both pretext and real questions in the testing prompts. The real question accuracy is presented in \cref{tab:aug_ablate}.
As expected, prior R1 models (Video-R1 and VideoRFT) exhibit significant performance drops due to perturbed video sequences and domain-shift. 
%
Our models demonstrate strong robustness to diverse video augmentations, with minimal accuracy loss. Finally, for our models, accuracy on the untransformed videos ($V$) is better than on transformed videos, which demonstrates the efficacy of the pretext task -- processing of the transformed videos in the pretext task encourages useful aggregation of information that transfers to untransformed videos.

\subsection{Training Curves}
We monitor key configuration dynamics during RL in \cref{fig:train_curve}.
The transformation reward $R_t$ in ViSS-R1 increases rapidly within the first 100 iterations and then stabilizes, indicating that the model quickly acquires pretext reasoning ability from the SFT stage.
%
Compared to vanilla GRPO only on user questions, our accuracy reward $R_a$ is noticeably lower (\cref{fig:train_curve}(b)), due to the increased difficulty introduced by visual transformations.
Meanwhile, as task difficulty increases, the \textit{all correct ratio} (the proportion of groups in which all generations are correct) decreases compared to vanilla GRPO. A higher all correct ratio suggests that more training samples are not truly learned (advantage is 0), which also explains why ViSS-R1 could outperform previous methods.
%
Furthermore, ViSS-R1 produces a longer average response (\cref{fig:train_curve}(d)), as it addresses two questions in a single forward pass.
The completion length initially drops and then stabilizes, reflecting that the model is discarding its SFT strategy and adapting to a new reasoning policy.

\subsection{Qualitative Result}
\label{subsec:vis}
We show qualitative results in \cref{fig:vis}.
%
Previous Video-R1 typically attempts to solve the problem directly and analyze the options (in MCQs), often without carefully reviewing the referenced video. However, visual reasoning is essential for correct inference in this context.
In contrast, our ViSS-R1, even without explicit instructions, exhibits a pattern of first analyzing the video content, then answering the question, and finally verifying the answer by checking available options.
The problem-solving solution aligns well with our initial objective: enabling visual-centric reasoning.
%
More examples can be found in the appendix.

\section{Conclusion}
\label{sec:conclusion}
We propose Pretext-GRPO, a pretext-based self-supervised reinforcement learning approach, along with its integration into the  R1-style framework, ViSS-R1, to address visual-centric complex video reasoning. 
By leveraging annotation-free SSL transformations, Pretext-GRPO enables thorough examination of visual content prior to text-based answering. 
The integrated ViSS-R1 further streamlines the training process and consistently improves overall video reasoning performance. 
We believe incorporating SSL mechanisms into MLLMs represents a promising direction for the development of future intelligent multimodal models.

{
    \small
    \bibliographystyle{ieeenat_fullname}
    \bibliography{main}
}


\end{document}